\documentclass[wcp]{jmlr}

\usepackage{longtable}

\usepackage{booktabs}

\usepackage{amsmath}

\usepackage{soul}
\usepackage{comment}

\newcommand{\expectancy}[3]{ \operatornamewithlimits{\mathbb{E}^{\:#3}}_{\:#1} \left [ #2 \right ] } 

\usepackage[]{algorithm}
\usepackage{algorithmic}

\newcommand{\G}{{\cal G}}

\makeatletter
\let\Ginclude@graphics\@org@Ginclude@graphics 
\makeatother

\jmlrvolume{}
\jmlryear{2022}
\jmlrworkshop{ACML 2022}

\title[When to Classify Events in Open Times Series?]{When to Classify Events in Open Times Series?}

  \author{\Name{Youssef Achenchabe} \Email{youssef.achenchabe@universite-paris-saclay.fr}\\
  \addr Paris-Saclay University, Agroparistech and Orange Labs.
  \AND
  \Name{Alexis Bondu} \Email{alexis.bondu@orange.com}\\
  \addr Orange Labs, Châtillon.
  \AND
  \Name{Antoine Cornuéjols} \Email{antoine.cornuejols@agroparistech.fr}\\
  \addr Paris-Saclay University, Agroparistech.
    \AND
  \Name{Vincent Lemaire} \Email{vincent.lemaire@orange.com}\\
  \addr Orange Labs, Lannion.
 }

\begin{document}

\maketitle

\begin{abstract}


In numerous applications, for instance in predictive maintenance, there is a pression to predict events ahead of time with as much accuracy as possible while not delaying the decision unduly. This translates in the optimization of a trade-off between earliness and accuracy of the decisions, 
that has been the subject of research for time series of finite length and with a unique label. And this has led to powerful algorithms for Early Classification of Time Series (ECTS).
This paper, for the first time, 
investigates such a 
trade-off when events of different classes occur in a streaming fashion,  with no predefined end.  
In the \textit{Early Classification in Open Time Series} problem (ECOTS), the task is to predict events, i.e. their class and time interval, at the moment that optimizes the accuracy vs. earliness trade-off.
Interestingly, we find that ECTS algorithms can be sensibly adapted in a principled way to this new problem.  
We illustrate our methodology by transforming two state-of-the-art ECTS algorithms for the ECOTS scenario.
Among the wide variety of applications that this new approach opens up, we develop here a predictive maintenance use case that optimizes alarm triggering times, thus demonstrating the power of this new approach. 
\end{abstract}

\section{Introduction}
\label{sec:intro}


In intensive care units \citep{shekhar2021benefit}, in control rooms of electrical power grids \citep{dachraoui2013early}, in government councils assessing emergency situations, in many kinds of contexts therefore, it is essential to make timely decisions in absence of complete knowledge of the true outcome. The issue facing the decision-makers is that, usually, the longer the decision is delayed, the clearer is the likely outcome (e.g. whether the patient is critical or not) but, also, the higher the cost that will be incurred if only because earlier decisions allow one to be better prepared. 
Formally, this problem translates into optimizing online the trade-off between the earliness and the accuracy of the decision. Early Classification of Time Series (ECTS) deals with time series of finite length, and a single decision per time series. 

In this paper, we study this trade-off in a new context, one where events of different classes occur in a streaming fashion, i.e. in a an open time series with no predefined end. We want to predict, at the optimal time, each of them with their associated time intervals.
For instance, in predictive maintenance, events may be associated with types of mechanical parts malfunctions that can be expected at given times.

Interestingly, we find that the powerful algorithms that have resulted from research on ECTS can sensibly be adapted to the new problem. 
In particular, this paper proposes a principled way to adapt ECTS approaches to deal with the early classification in open time series (i.e. with no time bounds). 
We show how the classification problem in ECTS can be transformed into a new classification problem for Early Classification in Open Times Series (ECOTS)  and how the decision triggering condition should be modified.



%
\smallskip
In the classical setting, the \textit{Early Classification of Time Series} (ECTS) problem assumes that measurements, possibly multivariate, become available over time in a time series which, at time $t$, is ${\mathbf x}_t \, = \, \langle {x_1}, \ldots, {x_t} \rangle \in {\mathcal X}^t$ where $x_t$ is the current measurement and each ${x_j}_{(1 \leq j \leq t)}$ belongs to some input domain $\mathcal X$ (e.g. the temperature and the blood pressure of a patient at time step $j$). 
It is further supposed that each time series can be ascribed to a class $y \in {\cal Y}$ (e.g. patient who needs a surgical operation or not). 
The task is to predict the class of an incoming time series by optimizing the trade-off between the expected accuracy of the prediction and the increasing cost of delaying the decision.

Formally, the ECTS problem can be stated as trying to find the best (possibly future) time $\tau^\star$ to make a prediction, given that only ${\mathbf x}_t$ has been observed:
\begin{equation}
    \tau^\star \; = \; \operatornamewithlimits{ArgMin}_{\tau \geq t} \; 
    \biggl\{\mathrm{C}_m(h_{\tau}(\mathbf{x}_\tau)| y) \, + \, \mathrm{C}_d(\tau) \biggr\}
\end{equation}
where $\mathrm{C}_m(h_{\tau}(\mathbf{x}_\tau)| y)$ is the cost of misclassification incurred at $\tau$, 
and $y$ is the true class, and  $\mathrm{C}_d(\tau)$ is the delay cost.
What makes the problem possible to tackle, in the ECTS scenario, is the availability of a training set $\mathcal S = \{({\mathbf x}_T^i, y_i)\}_{1 \leq i \leq m}$ of complete time series, with their associated labels. 
For each time step $t$, $t \in \{1, \ldots, T\}$, using $\mathcal S$, a classifier $h_t$ can be learned $h_t: {\cal X}^t \rightarrow {\cal Y}$, with ${\cal X}^t$ being the space of truncated time series at time step $t$.
These classifiers are learned beforehand and are considered as an input to the ECTS approaches, which essentially optimize the times at which predictions are triggered. 



\begin{figure}[htbp!]
\centering
\includegraphics[scale=0.5]{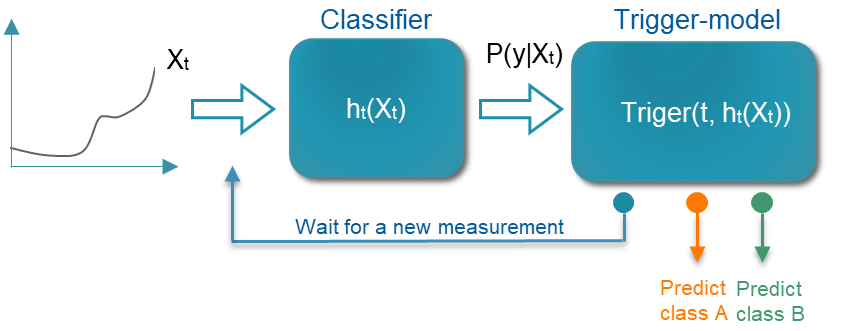}
\caption{General schema of ECTS approaches}
\label{fig-general-ECTS-shema}
\end{figure}

Overall, the ECTS problem can be seen as involving two components: (\textit{i}) the set of \textit{classifiers} $h_t$ ($1 \leq t \leq T$) which is supposed pre-learned as seen above, and (\textit{ii}) a \textit{ trigger system} which decides whether to make the prediction at the current time $t$ or to wait for at least one more measurement $x_{t+1}$ (see Figure \ref{fig-general-ECTS-shema}). 

\begin{figure}[htbp!]
\centering
\includegraphics[width=9cm,height=2.5cm]{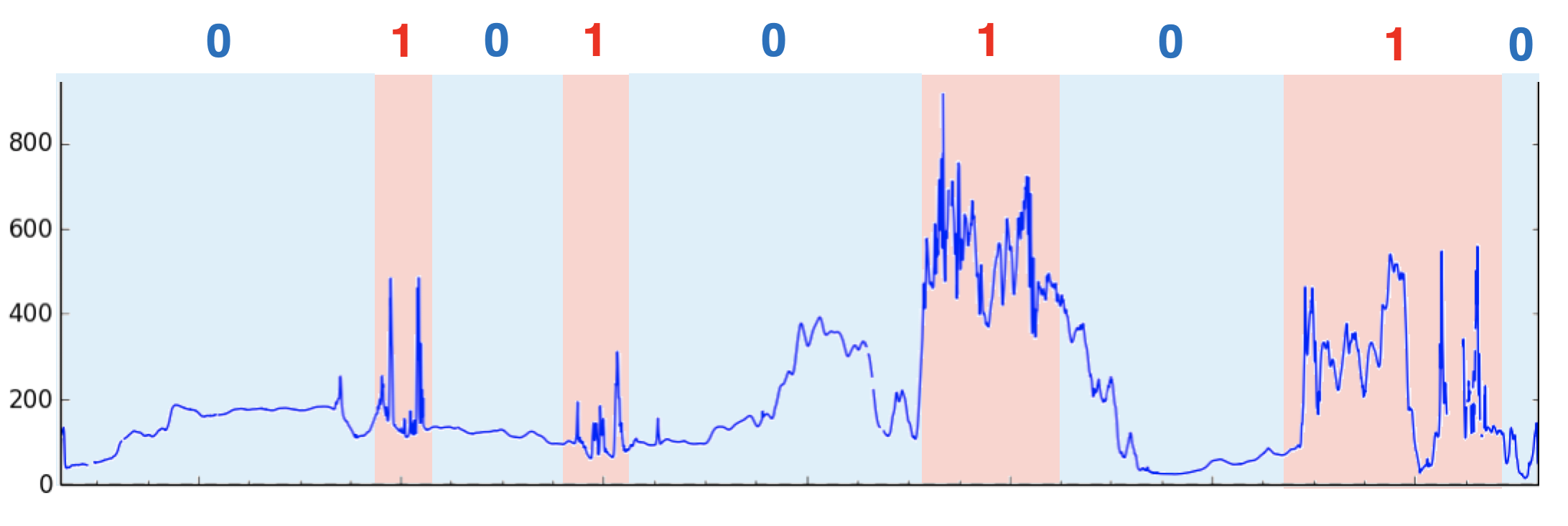}
\caption{Example of a part of an open time series where events of possibly different lengths are here labeled with `0s' and `1s'. }
\label{fig-labelled-stream}
\end{figure}

While the ECTS framework is well suited to many real world problems, it is limited in several ways, as has been underscored in \citep{bondu2022open} which defines several challenges that remain to be solved. In particular, one challenge relates to
the fact that, in ECTS,  (\textit{i}) the time series all have a finite length and (\textit{ii}) it is assumed that there is a single class associated with each of them. However, applications abound where the measurements come in an open time series with no time bounds and where different events may arise, possibly of different lengths, each with its own class (see Figure \ref{fig-labelled-stream}). 
The setting is thus different from ECTS. Should we then abandon the good properties and performances of the best ECTS algorithms for the new ECOTS problem?

\medskip
If not, adapting ECTS approaches to the ECOTS problem raises three issues:
\begin{enumerate}
    \item What should the \textit{classifiers} in ECOTS do? In the ECTS framework, they take incomplete time series ${\mathbf x}_t$ as input and make prediction about the class of the associated complete, but still unknown, time series ${\mathbf x}_T$. But, in ECOTS, the notion of complete time series does not make sense anymore.
    
    %
    \item How to solve the \textit{earliness-accuracy} trade-off in the ECOTS setting? In the ECTS framework, the start and length of time series are known which allows one to measure the earliness. But how to define it in ECOTS? 
    

    \item How to build the \textit{training sets} in the framework of ECOTS where there are no more individual time series with their associated class, but open time series with events of different classes and durations?
    
\end{enumerate}



The goal of this paper is, first, to define properly the ECOTS problem and, second, to present a methodology to adapt any ECTS approach to it by answering the three questions raised above. 
As a result, we show how the role of the classifiers must be thought anew and transformed and how the earliness-accuracy trade-off translates to the new scenario and what the decision triggering system becomes. 
%
%
We illustrate our methodology by transforming two state-of-the-art ECTS algorithms for the ECOTS scenario.
Among the wide variety of applications that this new approach opens up, we develop here a predictive maintenance use case that optimizes alarm triggering times, thus demonstrating the power of this new approach.

In order to avoid confusion, it is important to note that the data stream literature \citep{silva2013data} focuses on classification of incoming data points at fixed horizon under memory constraints and evolving properties of data. Whereas, in this paper, we focus on identifying the optimal moment of the classification, the one that optimizes the \textit{earliness-accuracy} trade-off in a stationary environment. 

The paper is organized as follows. Section \ref{sec_ects_ecots} formally draws a parallel between the ECTS problem and the ECOTS one, leading to a generic approach capable of transposing any ECTS algorithm into an ECOTS one. We then review, in Section \ref{sec_related_work}, the main approaches to ECTS, and outline two competitive methods: one described in \citep{mori2017reliable}  and the other, the \textsc{Economy} method presented in \citep{achenchabe2021earlyy}. We then show, in Section \ref{sec_adapted_methods}, how to adapt these methods to the ECOTS problem, before comparing their performances on experiments in Section \ref{sec-experiments}. The conclusion, in Section \ref{sec-conclusion}, underlines what has been performed in this work, and provides directions for future research.

\section{ECOTS in the perspective of ECTS}
\label{sec_ects_ecots}
This section defines the ECTS and ECOTS problems in turn, and presents our proposed methodology to adapt any ECTS approach to solve the ECOTS problem.

\subsection{The ECTS problem}
\label{sec_ects}

\begin{figure}[htbp!]
\centering
\includegraphics[width=5cm,height=1.5cm]{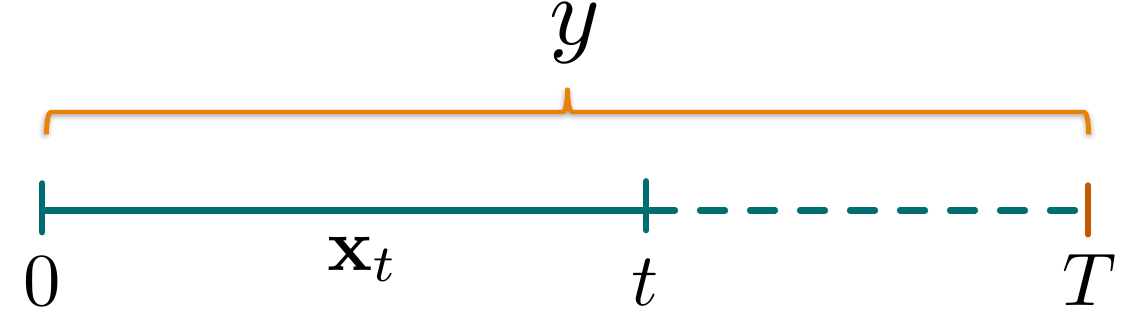}
\caption{In the ECTS setting, the classifier $h_t$ sees the incoming time series ${\mathbf x}_t$ and predicts a label $\hat{y}$ of the complete time series ${\mathbf x}_T$ of true class $y$.}
\label{fig-ects-1}
\end{figure}

In the ECTS setting, each classifier $h_t$ ($1 \leq t \leq T$) is learned from the truncated training time series up to time $t$: $\{({\mathbf x}_t^j, y_j)\}_{(1 \leq j \leq m)}$ 
(see Figure \ref{fig-ects-1}). It is expected that the accuracy of the classifiers grows as $t$ increases from the first time step $t=1$ to the last one $t=T$.  

The problem, given an incoming time series, is to choose a time $t$ for which the expected cost of misclassification $\mathrm{C}_m(\hat{y}|y)$, where $\hat{y} = h_t({\mathbf x}_t)$ and $y$ is the true class, plus the delay cost $\mathrm{C}_d(t)$, is minimal. Formally, the combined expected cost is given by:


\begin{equation}
 \begin{split}
        \normalsize
        f(\mathbf{x}_t) \; &= \; \expectancy{(\hat{y},y) \in  {\cal Y}^2}{\mathrm{C}_m(\hat{y}| y)|\mathbf{x}_t}{t} \; +  \mathrm{C}_d(t)  \\
        &= \; \sum_{y \in {\cal Y}} P_{t}(y|\mathbf{x}_t) \, \sum_{\hat{y} \in {\cal Y}} P_t(\hat{y}|y, \mathbf{x}_t) \, \mathrm{C}_m(\hat{y}|y) \; + \; \mathrm{C}_d(t)
        \label{eq:cost1}
\end{split}    
\end{equation}
where $\expectancy{(\hat{y},y) \in  {\cal Y}^2}{}{t}$  is the expectancy at time $t$, over the variables $y$ and $\hat{y}$. $P_{t}(y|\mathbf{x}_t)$ is the probability of the class $y$ given an incomplete time series ${\mathbf x}_t$, and $P_t(\hat{y}|y, \mathbf{x}_t)$ is the probability that the classifier $h_t$ makes the prediction $\hat{y}$ given ${\mathbf x}_t$ as input while $y$ would be its true label. 

The objective of the trigger function is to identify the best time $t^\star$ for triggering the decision while receiving online the measurements of the time series, the one that minimizes Equation \ref{eq:cost1}. Section \ref{sec_related_work} provides a state of the art of the existing approaches to solve this problem, from very heuristic ones to more formally grounded.

\subsection{The ECOTS problem}
\label{sec_ecots}

In this paper, contiguous instants with the same label are called chunks or events (see Figure \ref{fig-labelled-stream}).
In the ECOTS scenario, we suppose that we have a training data set of $m$ labeled chunks $\{({\mathbf x}^j, y_j)\}_{(1 \leq j \leq m)}$ coming from an open time series, where each ${\mathbf x}^j$ has length $l_{\mathbf{x}^j}$, and where $y_j$ is the corresponding label. The ECOTS problem consists in predicting as soon as possible these events, i.e. their class label and time limits.

\subsection{Proposed transposition of ECTS approaches into ECOTS ones}
\label{sec_methodo}

We propose to simplify the ECOTS problem by considering point-wise predictions instead of predicting whole chunks (i.e. class and time limits), that is to make independent predictions of the labels associated with each single timestamp $t_p$ in the open time series. From such predictions, chunks could be reconstituted, for instance by gluing time stamps with the same predicted class.

We posit that the unfolding time series is observed over a finite time interval which depends on the present time $t$. 
In the following, we will use the term time window and note it ${\mathbf x}_{(t-w, t)}$ at time $t$ when its size is $w$ ($w$ past measurements are available at time $t$). 

Then, the target time stamp $t_p$ can be in the future ($t_p > t$), for example if there are warning signs that a machine will break down, or in the past ($t_p < t$) if it was necessary to wait until $t_p$ was in the observation time window to be able to identify its class, as it can happen during a computer attack for example.
Therefore, at any time $t$, for a target timestamp $t_p$, the transposed ECOTS problem is to decide whether $t$ is the best time for the prediction $\hat{y}_{t_p}$, the class associated with $t_p$,  or whether to postpone this decision to the next time step $t+1$, which will bring a new measurement.

\begin{figure}[htbp!]
\centering
\includegraphics[width=9cm]{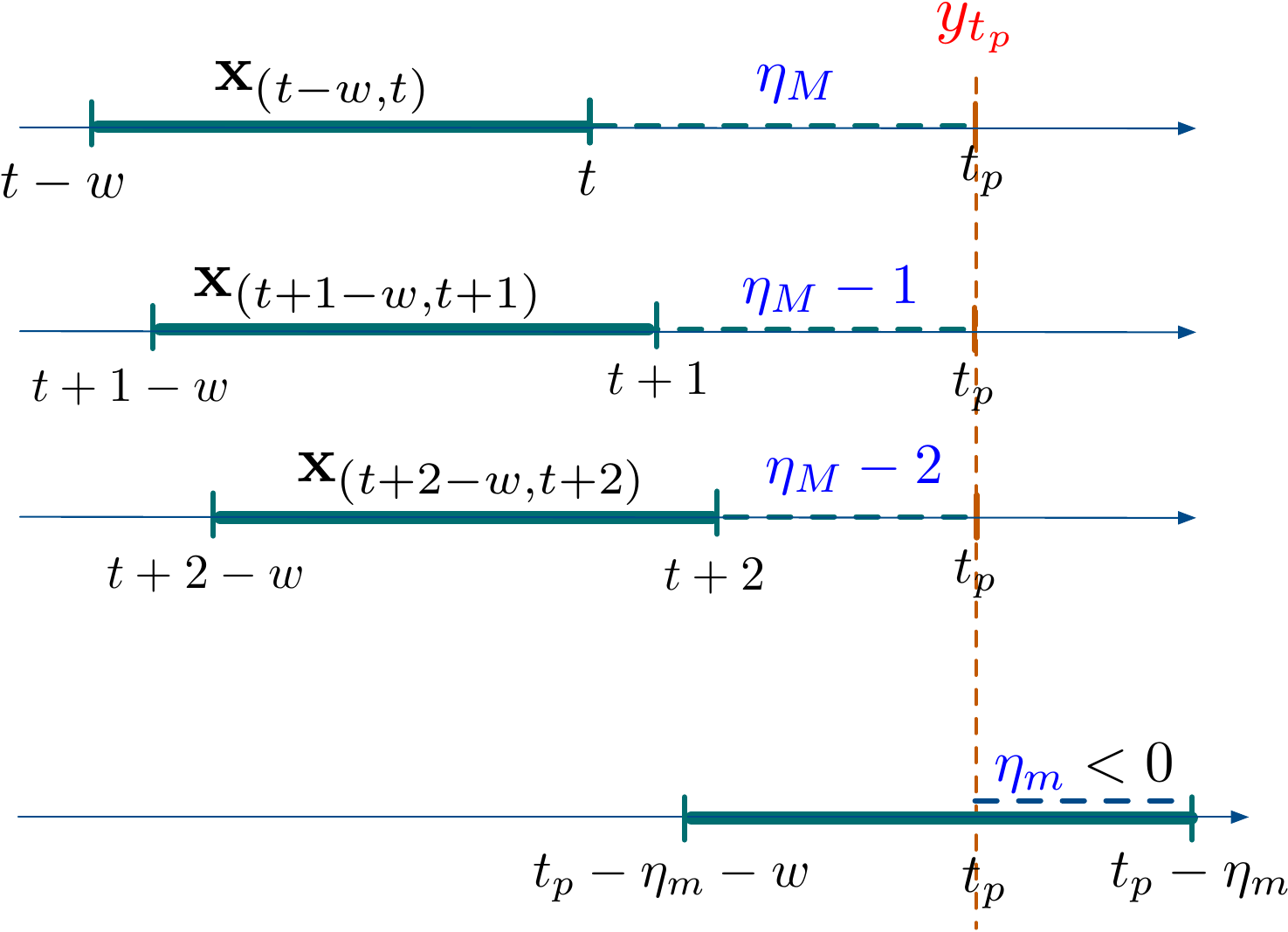}
\caption{The fixed point in time where to make a prediction is $t_p$, of true label $y_{t_p}$. As the measurements become available from $t_p-\eta_M$ to $t_p-\eta_m$, different classifiers $h_\eta$ come into play with an advancing sliding window and a diminishing horizon. The triggering system selects the best time to make a prediction $\hat{y}_{t_p}$.}
\label{fig-ecots-2}
\end{figure}

It is expected that, as the window of observation ${\mathbf x}_{(t-w, t)}$ comes closer to $t_p$, with increasing $t$, it is easier to make a reliable prediction about its class, but, at the same time, the cost of delaying prediction increases. (See Figure \ref{fig-ecots-2}). Note that we assume that there is a maximal value $\eta_M$ for the horizon, above which no precursor signal can be detected, and a minimal one $\eta_m$ after which it is no longer useful to detect the event.

%
%
 %


 

\smallskip
We now turn to the three questions raised in the introduction.


\smallskip
\noindent
\textbf{1}- In the classical ECTS problem (see Section \ref{sec_ects} and Figure \ref{fig-ects-1}), each classifier $h_t$ observes a time series $\mathbf{x}_t$ that is increasingly large as $t$ approaches the time limit $T$. In the ECOTS setting, as we propose to see it, by contrast, each classifier $h_\eta$ observes a sliding window $\mathbf{x}_{(t-w, t)}$ of the same number $w$ of observations. And each one makes a prediction about the class of the time $t_p$ positioned at a given horizon $\eta$ (positive or negative) from $t$ (see Figure \ref{fig-ecots-2}). Given a maximal horizon $\eta_M > 0$, the first classifier that can make a prediction about a time $t_p$ is $h_{\eta_M}$ viewing the window $\mathbf{x}_{(t_p-\eta_M-w, t_p-\eta_M)}$. And the last classifier is $h_{\eta_m}$ viewing the window $\mathbf{x}_{(t_p-\eta_m-w, t_p-\eta_m)}$. 
Thus, instead of having a set of classifiers $h_t$ ($1 \leq t \leq T$) {as in the ECTS setting}, we now have a set of classifiers $h_\eta$ ($\eta_m \leq \eta \leq \eta_M$) with various horizons. (See Figure \ref{fig-ecots-2}).


\medskip
\noindent
\textbf{2}- The second question to solve is to adapt the earliness-accuracy trade-off to the ECOTS problem. 
In the proposed transposition, the gain of information over time is due to a time window $\mathbf{x}_{(t-w, t)}$ that gets closer to the target timestamp $t_p$, and no longer to  additional measurement gathered over time as in ECTS. In addition, the single prediction triggered for each time series in ECTS is replaced by a prediction for each possible target timestamp in the open time series. These two essential but easy to carry out modifications allow to transpose most of the ECTS approaches, since the optimization criterion they use for their triggering strategy can remain essentially the same. 
%
%
%
As a proof of concept, in Section \ref{sec_adapted_methods}, we show how two state of the art methods for ECTS can be adapted to the ECOTS problem.

\medskip
\noindent
\textbf{3}- In the proposed approach, the values of the window size $w$, and of the bounds $\eta_m$ and $\eta_M$ of the horizon have to be chosen from a training set. But which training set?

\smallskip
In the ECOTS problem, it is assumed that the relationship between the symptoms of an event and the event itself are \textit{stationary} over time (e.g. a given malfunction of a machine keeps the same telltale signs and the same characteristics whenever its appearance in time. Or the symptoms associated with a patient who should undergo an urgent heart operation stay the same, fortunately for the doctors, and for their training). 
From this property, in the same way as doctors can be trained using independent episodes in an hospital history about heart attacks, 
it is possible to use subsequences of the open time series, as long as they are independent, to build training datasets in order to learn the classifiers $h_\eta$ (see Section \ref{sec:exp_protocol}).



\section{Related work on trigger systems}
\label{sec_related_work}


While most of the proposed ECTS approaches rely on learning classifiers for different time steps in the time series, they differ in the way they trigger the decision to make a prediction.

Two different approaches can then be used. In the \textit{myopic} one, at each successive time step $t$, the accuracy (or confidence) of the current classifier $h_t$ on its prediction can be assessed and it is possible to decide whether the time seems right to make a prediction. In a \textit{non-myopic} approach, given the current incoming time series ${\mathbf x}_t$, the expected accuracy of all future classifiers $h_\tau$ ($t \leq \tau \leq T$) given the current input can be estimated and the best decision time be computed taking into account the delay cost. Only if $t$ coincides with this best instant, the prediction is performed

The earliest works on ECTS were heuristics in nature as well as myopic. They did not try to explicitly optimize the cost defined in equation (\ref{eq:cost1}) but instead relied on the concept of an estimated confidence in the current prediction. If this estimate was above a threshold, then the decision was triggered. For instance, \citep{mori2017reliable} describes a method where the accuracy of a set of probabilistic classifiers is monitored over time, which allows the identification of time steps from whence it seems safe to make predictions. In \citep{parrish2013classifying,hatami2013classifiers,ghalwash2012early}, various stopping rules are defined, some on them relying on a confidence level threshold. And in \citep{xing2009early}, the best time step to trigger the decision is estimated by determining the earliest time step for which the predicted label does not change, based on a 1-NN classifier.

More recently, approaches have been proposed that attempt to explicitly optimize the trade-off between the earliness and the accuracy. A notable example is \citep{mori2017early} where a single objective optimization criterion is defined. In \citep{mori2019early}, the authors put forward the idea of a multi-objective criterion with an associated Pareto front with multiple dominating trade-offs.
However, whereas these methods are \textit{myopic} in nature, the \textsc{Economy} approach described in \citep{achenchabe2021earlyy,dachraoui2015early} goes one step further by (\textit{i}) directly optimizing the combined cost defined in equation \ref{eq:cost1} and (\textit{ii}) giving a way to estimate the combined cost for future time steps, thus leading to a \textit{non-myopic} approach that outperforms the best methods known to date.

For a recent survey, the interested reader can refer to \citep{gupta2020approaches}.


\section{Adapting two state-of-the art ECTS approaches}
\label{sec_adapted_methods}


In Section \ref{sec_methodo}, we have shown how to translate an ECTS problem into an ECOTS one by modifying the definition and purpose of \textit{the classifiers}.
In the following, we demonstrate how the \textit{triggering strategy} used in ECTS can be adapted to deal with ECOTS. For this, we consider one of the best performing myopic strategies known to date, described in \citep{mori2017early} and the best non-myopic approach in the literature: the \textsc{Economy}-$\gamma$ strategy described in \citep{achenchabe2021earlyy}. The first one relies ultimately on confidence criteria, while the second one explicitly optimizes the accuracy versus delay cost trade-off.

\subsection{The SR approach}
\label{sec_SR_approach}
The \textsc{SR} approach \citep{mori2017early} 
uses a trigger-model which involves 3 parameters $(\gamma_1, \gamma_2, \gamma_3)$ in order to decide if the current prediction is reliable (output $1$) or if it is preferable to wait for more data (output $0$):

\begin{equation}
\textit{Trigger}^\gamma \left ( h_\eta({\mathbf x}_t) \right )  =  
    \begin{cases}
        0 & \mbox{if } \gamma_1 \, p_1 + \gamma_2 \, p_2 + \gamma_3 \, \frac{t}{T} \le 0 \\
        1 & \mbox{otherwise}
    \end{cases}
\label{eq:mori_trigger_function}
\end{equation}
where $p_1$ is the largest posterior probability estimated by the classifier $h_\eta$, $p_2$ is the difference between the two largest posterior probabilities, and the last term $\frac{t}{T}$ represents the proportion of the incoming time series that is visible at time $t$.  
The parameters $\gamma_1, \gamma_2, \gamma_3$ are real values in $[-1, 1]$ to be optimized using training data. 

\begin{algorithm}[h]
\begin{algorithmic}[1]
\small
    \REQUIRE
     $t$: current moment.\\
     $t_p$: target that belongs to [t+$\eta_m$, t+$\eta_M$].\\
     $w$, $\eta_m$, $\eta_M$: window size, minimum and maximum horizon.\\

 	\FORALL{$ \eta = t_p-t, \dots, \eta_m$ (step=-1)} 
 	 \STATE sliding\_window = $\mathbf{x}_{(t_p-\eta-w, t_p-\eta)}$ \# sliding window: adds new measurement and deletes the first one.\\
	\STATE compute A = $\gamma_1 \, p_1 + \gamma_2 \, p_2 + \gamma_3 \, \frac{\eta_M -\eta}{\eta_M-\eta_m} $ using the updated sliding window.\\
	
	\IF{$A > 0$ or $\eta == \eta_m$}
	    \STATE return $\eta$
	\ENDIF
	\ENDFOR

\end{algorithmic}
\caption{\small Adapted \textsc{SR} approach in the ECOTS scenario}
\label{algo:learn_mori}
\end{algorithm}

In the ECOTS problem, the trigger function for a target at horizon $\eta = t_p - t$ becomes:
\begin{equation*}
\small
\textit{Trigger}^\gamma \left ( h_\eta({\mathbf x}_{(t-w, t)} \right )  = 
    \begin{cases}
        0 & \text{if } \gamma_1 \, p_1 + \gamma_2 \, p_2 + \gamma_3 \, \frac{\eta_M -\eta}{\eta_M-\eta_m} \le 0 \\
        1 & \text{otherwise}
    \end{cases}
\label{eq:mori_open}
\end{equation*}

\textcolor{red}{}

The last term of Equation (\ref{eq:mori_trigger_function}) is replaced by $\frac{\eta_M -\eta}{\eta_M-\eta_m}$, which represents the relative position of the current horizon $\eta$ in the considered range of horizons $[\eta_m,\eta_M]$. Algorithm \ref{algo:learn_mori} highlights the adapted procedure of the SR approach to choose the optimal horizon for a given target $t_p$.




\subsection{The ECONOMY-$\gamma$ approach} 
\label{sec:eco_approach}




\textsc{Economy}-$\gamma$ is a non-myopic cost-based approach \citep{achenchabe2021early}, which is capable of estimating the expected cost of making a prediction for any time $t + \tau$ ($1 \leq \tau \leq T-t$) in the future, defined as:  
\begin{equation}
 \begin{split}
        \normalsize
        f_{\tau}(\mathbf{x}_t)&= \expectancy{(\hat{y},y) \in  {\cal Y}^2}{\mathrm{C}_m(\hat{y}|y) | \mathbf{x}_t}{t+\tau} \; +  \; \mathrm{C}_d(t+\tau) \\ 
        &= \; \sum_{y \in {\cal Y}} P_{t+\tau}(y|\mathbf{x}_t) \, \sum_{\hat{y} \in {\cal Y}} P_{t+\tau}(\hat{y}|y, \mathbf{x}_t) \, \mathrm{C}_m(\hat{y}|y) \; + \; \mathrm{C}_d(t+\tau)
        \label{eq:cost2}
 \end{split}  
\end{equation}
\noindent

In practice, the terms $P_{t+\tau}(\hat{y}|y, \mathbf{x}_t)$ and $P_{t+\tau}(y|\mathbf{x}_t)$ are not tractable. A partitioning of the training data into $K$ groups $\mathfrak{g}_k \in \mathbf{G}$ (see \citep{achenchabe2021early})
%
is required to make them computable, yielding the following approximation:
\begin{equation*}
 \begin{split}
        \normalsize
        f_{\tau}(\mathbf{x}_t) \approx \sum_{\mathfrak{g}_k \in \G} P_{t+\tau}(\mathfrak{g}_k|\mathbf{x}_t) \sum_{y \in {\cal Y}} P_{t+\tau}(y|\mathfrak{g}_k) \, \sum_{\hat{y} \in {\cal Y}} P_{t+\tau}(\hat{y}|y, \mathfrak{g}_k) \, \mathrm{C}_m(\hat{y}|y) \; + \; \mathrm{C}_d(t+\tau)
        \label{eq:cost-grp}
 \end{split}  
\end{equation*}


The optimal decision time, at time $t$, is thus estimated to be:
\begin{equation}
\label{eq:time2}
\tau^\star \; = \; \operatornamewithlimits{ArgMin}_{\tau \in \{0, \ldots, T-t\}} f_\tau(\mathbf{x}_t)
\end{equation}

The idea is to estimate the cost of a decision for all future time steps, up until  $t=T$, based on the current knowledge about the incoming time series $\mathbf{x}_t$. The decision is postponed unless $\tau^* = 0$, that is when it is expected that there will be no better trade-off in the future.  If so, the prediction $h_t({\mathbf x}_t)$ is returned and the classification process is terminated. Otherwise, the decision is postponed to the next time step, and Eq. (\ref{eq:time2}) is computed again, this time with ${\mathbf x}_{t+1}$ as input. The process goes on until a decision is made or $t = T$ at which point a prediction is forced.

\begin{algorithm}[h]
\begin{algorithmic}[1]
{\small
    \REQUIRE     $t$: current moment.\\
     $t_p$: target that belongs to [t+$\eta_m$, t+$\eta_M$].\\
     $w$, $\eta_m$, $\eta_M$: window size, minimum and maximum horizon.\\

 	\FORALL{$\eta = t_p-t, \dots, \eta_m$} 
		\STATE compute $f_{\eta}({\mathbf x}_{(t-w+1, t)})$
	\ENDFOR	
	\STATE $\eta^\star \; = \; \operatornamewithlimits{ArgMin}_{\eta_m \leq \eta \leq t_p-t} f_{\eta}({\mathbf x}_{(t-w+1, t)})$
	\IF{$\eta^\star$ == $t_p$-t or $t_p-t == \eta_m$}
	    \STATE return $\eta^\star$
	\ENDIF

	\STATE return None
	
}
\end{algorithmic}
\caption{\small Adapted \textsc{economy} approach in the ECOTS scenario}
\label{algo:learn_ECO_ecots}
\end{algorithm}

\medskip
In ECOTS, as time $t$ increases, the task is to label each time step $t_p$ as it appears in the span of the horizon: [$t+\eta_m, t+\eta_M$] (see Figure \ref{fig-ecots-2}). 
While in Equation (\ref{eq:cost2}), the term $\expectancy{(\hat{y},y) \in  {\cal Y}^2}{\mathrm{C}_m(\hat{y}|y)}{t+\tau}$ involves the calculation of the confusion matrices for future time steps $t+\tau$ knowing the current incoming time series ${\mathbf x}_t$, the adaptation to ECOTS requires that the confusion matrices are now computed for the various horizons from $\eta_m$ to $\eta_M$ and then used to estimate the cost of decision for each horizon $\eta$: 
\begin{equation*}
 \begin{split}
        f_{\eta}({\mathbf x}_{(t-w, t)}) \; = \; \expectancy{(\hat{y},y) \in  {\cal Y}^2}{\mathrm{C}_m(\hat{y}|y) | {\mathbf x}_{(t-w, t)}}{\eta} \; +  \; \mathrm{C}_d(\eta) \;  
 \label{eq:cost3}
 \end{split}    
\end{equation*}
\noindent
and the best horizon:
\begin{equation*}
    \eta^\star \; = \; \operatornamewithlimits{ArgMin}_{\eta_m \leq \eta \leq t_p-t} f_{\eta}({\mathbf x}_{(t-w, t)})
\end{equation*}

The decision to classify the data point $t_p$ is triggered at time $t$ either because $t+\eta^\star = t_p$ (i.e. corresponding to the optimal cost) or when $t_p = t+ \eta_m$ (i.e. it is not possible to wait any longer), see Algorithm \ref{algo:learn_ECO_ecots} for more details.

\smallskip
The cost of delay $\mathrm{C}_d(t)$, which is an \textit{increasing} function of $t$ in ECTS is a \textit{decreasing} function $\mathrm{C}_d(\eta)$ of the horizon $\eta$ in ECOTS. Indeed, as the target that we want to label approaches ($\eta$ decreasing), the cost of the decision increases. 

\smallskip
Note that the time and space complexities of the \textsc{Economy}-$\gamma$ approach adapted to ECOTS are the same than in the original approach. In the ECTS setting, at testing phase, computing the cost at each timestep is in O($T^2$) at worst case, and in ECOTS setting is in $O((\eta_M-\eta_m)^2)$.


\section{Experiments}
\label{sec-experiments}


We have proposed a principled method to adapt any ECTS approach into an ECOTS one (see Sections \ref{sec_ects_ecots} and \ref{sec_adapted_methods}). 
The aim of the experiments is to validate that the adaptation of the SR and \textsc{Economy}-$\gamma$ approaches is efficient in the ECOTS setting. 
In addition, we illustrate the applicability of the proposed approaches for predictive maintenance using real data from the industrial domain. 

%
\smallskip
This section aims at answering the following questions:
\begin{enumerate}
\item How efficient is the proposed framework for adapting any ECTS approach to the ECOTS problem compared to baseline algorithms designed for the ECOTS problem?
\item How do these approaches behave when the delay cost increases, and when the misclassification cost becomes very imbalanced?
\item How these approaches adapt their decision time to the observed data? 
\end{enumerate}

Our \textit{source code} is shared for full reproducibility of the experiments in the supplementary material. This also allows interested researchers to extend the experiments to other open time series datasets. 

\subsection{Experimental protocol}
\label{sec:exp_protocol}

\subsubsection{Data description:}
We use an open real dataset \citep{data} from one of the Schwan’s factories. It contains 100 multivariate time series corresponding to 100 machines monitored over time for a period of 1 year (January 2015 to January 2016) with measurements collected every hour. Each time series is a multi-dimensional data table whose rows indicate the temporal domain and columns include telemetry features (pressure, rotation, voltage and vibration), 5 Boolean columns encoding different types of device errors which are premises correlated with a future failure and a last column which indicates the presence or absence of a failure (the variable to be predicted). This makes 8761 rows and 10 columns for each machine. The whole dataset contains 3919 errors and 761 failures for a total number 876,100 timestamps. This dataset is extremely imbalanced with 0.08\% of timestamps associated with the abnormal class (i.e. failure). There is on average 7 failures per time series, with a minimum of 0 and a maximum of 19 failures per machine during the observed year.

\subsubsection{Problem statement:}
Traditionally, the problem of predictive maintenance is solved by fixing a horizon for predictions (e.g. if a technician needs at least 12 hours to take preventive actions before the machine fails actually, then a fixed horizon would be chosen as $\eta = $12 hours). Our goal is to use the ECOTS approaches to make this horizon \textit{adaptive} to the observable part of the time series at hand. 


\subsubsection{Evaluation criterion:}
Ultimately, the value of using an early classification method is defined by the \textit{average cost} incurred using it, as in \citep{achenchabe2021earlyy}. Given an open time series ${\cal S}$ (e.g. a machine monitored over a year), observed on a finite time interval of sufficient length $N$. This time period is composed of time stamps $t \in [1,N]$, labeled by the class $y_t$.
As time increases from 1 to $N$, the ECOTS system makes predictions for each time step $t$: $\hat{y}_t$ while the true class is $y_t$. In addition, for each $t \in [1,N]$, the prediction is made using a classifier $h_{\eta_{t^\star}}$ corresponding to the triggering horizon $\eta_{t^\star}$,   
 thus incurring a delay cost $\mathrm{C}_d(\eta_{t^\star})$. Hence, the formula:
\begin{equation}
\begin{split}
\textit{AvgCost}({\cal S}) \;  =  \frac{1}{N} \sum_{t=1}^{N}  \bigl( \mathrm{C}_m ( \hat{y}_{t}|y_t)  \; + \; \mathrm{C}_d(\eta_{t^\star}) \bigr)   
\end{split}
\label{eq:eval-cost}
\end{equation}

\subsubsection{Computing the costs in the experiments:} In real applications, the decision costs would be provided by domain experts. In order to study the behavior of the different ECOTS algorithms, a large range of values has been considered for the misclassification and the delay costs.


\medskip
\noindent
\textit{The cost of misclassification}: Since we deal with a predictive maintenance problem, we make the assumption that the cost of missing a failure is much higher than the cost of sending the technical team. We thus consider four different misclassification costs $C_m = $
$\begin{bmatrix}
TN & FN \\
FP & TP 
\end{bmatrix}$, by varying the importance of false negatives: \\
~\\
$C_m^{(1)} = $
$\begin{bmatrix}
0 & 1 \\
1 & 0 
\end{bmatrix}$
, $C_m^{(2)} = $
$\begin{bmatrix}
0 & 10 \\
1 & 0 
\end{bmatrix}$
, $C_m^{(3)} = $
$\begin{bmatrix}
0 & 100 \\
1 & 0 
\end{bmatrix}$
, $C_m^{(4)} = $
$\begin{bmatrix}
0 & 1000 \\
1 & 0 
\end{bmatrix}$.\\

\medskip
\hspace{-0.56cm}\textit{The cost of delaying decision}: 
The delay cost $\mathrm{C}_d(\eta)$ is provided by the domain experts for an actual use case, and could be of any form. In our experiments, we set it as a linear function of horizon, with coefficient, or slope, $\alpha$: $\mathrm{C}_d(\eta) = \alpha \times \frac{\eta_M-\eta}{\eta_M - \eta_m}$. 
The larger $\alpha$, the higher the cost of postponing the decision and the greater the incentive to make prediction for large horizons  $\eta$. 
When $\alpha$ is very high, the gain in misclassification cost by waiting to be closer to the target cannot compensate for the increase of the delay cost, and it is better to make a decision early on, that is for large horizons, close to $\eta_M$.  
If, on the contrary, $\alpha$ is very low compared to the misclassification cost, it does not hurt to wait until the target $t_p$ is close to the sliding window ${\mathbf x}_{(t_p-\eta-w, t_p-\eta)}$. 
Our experiments were run over a large range of values of $\alpha \in$[10e-04, 10e-03, 10e-02, 1, 10, 100, 1000].

\subsubsection{Training the collection of classifiers and ECOTS algorithms}
\noindent
\textit{\hspace{-0.0cm}Data split and extraction:}
We splitted the set of time series into four parts: 50\% for training the classifiers, 20\% for testing the ECOTS algorithms, 15\% for validating the ECOTS algorithms and 15\% for estimating the confusion matrices. This split is inspired from the original paper of \textsc{economy} \citep{achenchabe2021earlyy}.  Subsequences of size $w$ were extracted from the training open time series by doing the following steps: (\textit{i}) time stamps $t_p$, aka targets, were set within the time series, spaced with $w+\eta_M$ time units in order to avoid overlaps between samples; (\textit{ii}) $\eta_M - \eta_m$ subsequences of size $w$ were extracted around each target, each one dedicated to the training of the classifier $h_{\eta}$ (see Figure \ref{fig-ecots-2}).

\smallskip
\noindent
\textit{Choice of the parameters $w$, $\eta_m$, $\eta_M$: } These parameters depend on the problem that is being solved and the data associated with it. One of the key ingredients of early classification methods is the information gain measured by the AUC. Generally, the expected cost of misclassification decreases as the target being classified gets closer to the sliding window. 
A window size of $w=10$ has been chosen to study the behavior of the ECOTS problems, since it shows a significant information gain over various horizons using AUC. We refer the reader to the supplementary material for AUC curves as a function of horizon with different sliding window sizes. They exhibit equivalent information gain curves, which means that this dataset is not very sensitive to the choice of $w$.
The parameter $\eta_M$ can be chosen according to the AUC, for our experiments we have chosen $\eta_M = 50$ as the AUC reaches 0.5 which corresponds to the random model, while, for $\eta_m$, we chose the end of the sliding window: $\eta_m = -w$.

\smallskip
\noindent
\textit{Training the collection of classifiers: }
As mentioned in Section \ref{sec_ecots}, a set of classifiers $h_\eta$ for different horizons $\eta$ such that ($\eta_m \leq \eta \leq \eta_M$) has to be trained. Extracted features\footnote{Reproducible using our source code available in the supplementary material.} from sliding windows include simple statistics: \textit{min}, \textit{max}, \textit{mean}, \textit{median} and the \textit{count} of each type of errors. For our experiments, we trained XGboost models by fine tuning parameters within the following grid of values\footnote{The interested reader can refer to the official documentation for more details: \url{https://xgboost.readthedocs.io/en/stable/}}: 
\textit{min chlid weight} $\in [1, 5, 10]$,
\textit{gamma} $\in [0.5, 1, 1.5, 2, 5]$,
\textit{subsample} $\in [0.6,0.8,1]$,
\textit{colsample} by tree $\in [0.6, 0.8, 1]$,
\textit{max depth} $\in [3,4,5,10]$.
The parameter \textit{scalePosWeight} is set to the the ratio of positive examples over negative ones in order to take into account the fact that the dataset is imbalanced.  
The combination of parameters that minimize the total cost\footnote{Given the cost of false positives and false negatives, the total cost is computed on a validation set as the sum over wrongly predicted samples weighted by the corresponding cost.} is chosen on a validation set (20\% of the set used for training the classifiers), then the model is learned on the whole training set. 
For a fair comparison, the same collection of classifiers is used for all ECOTS algorithms.


\noindent
\newline
\textit{ECOTS algorithms: }
The competing approaches considered in this paper are described below as well as their optimized hyper-parameters.
\begin{itemize}
    \item \textcolor{gray}{\bf Late baseline}: the last classifier in the collection $\Tilde{h}_{\eta_m}$ is used. This is the last time that a prediction can be made. 
    \item \textcolor{blue}{\bf Early baseline}: the first classifier in the collection $\Tilde{h}_{\eta_M}$ is used. This corresponds to the earliest possible prediction with the largest horizon in the future. 
    \item \textcolor{black}{\bf Confidence-based Classifiers (CC)}: for a fixed target $t_p$, this method takes a decision as soon as the confidence of the classifier regarding the class of interest exceeds a given threshold, optimized as a meta-parameter for each value of $\alpha$ using validation set. If this never happens, then $t_p = t+\eta_m$ and the prediction is forced using $\Tilde{h}_{\eta_m}$. 
    \item \textcolor{red}{\bf Economy-$\gamma$} (see Section \ref{sec:eco_approach}): for each value of $\alpha$, the number of groups $K$ used in the method is optimized in the range [1,5] using a validation set.
    \item \textcolor{orange}{\bf SR} (see Section \ref{sec_SR_approach}): for each value of $\alpha$, the parameters $\gamma_1$, $\gamma_2$ and $\gamma_3$ were optimized in the range $[-1, -0.5, 0,$ $0.5, 1]^3$ using a validation set.
\end{itemize}
Note that the ``late'' and the ``early'' baselines are not adaptive, while the ``Confidence-based'' method adapts its decisions to the current input. One goal of the experiments is to compare these methods with ones that have been \textit{translated} from the ECTS framework: {\bf Economy-$\gamma$} and {\bf SR}.

\subsection{Results and analysis} 


In this section, detailed answers to the questions raised in the introduction of Section \ref{sec-experiments} are given, supported by numerical results.

\begin{figure}[h]
\centering
\includegraphics[scale=0.74]{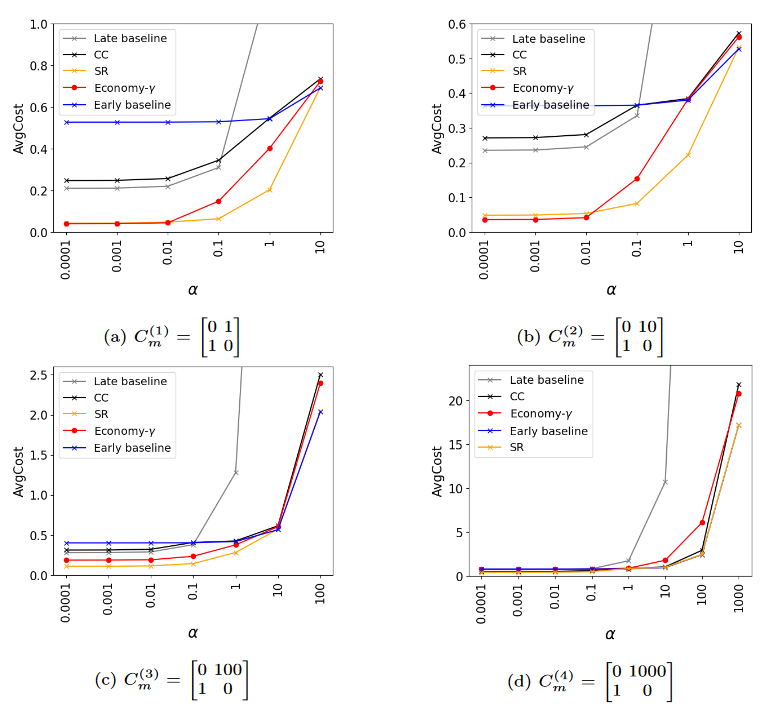}
\caption{\textit{AvgCost} of ECOTS algorithms computed on the test set for different values of the $\alpha$ parameter of the delay cost (x-axis), and for different values of misclassification cost. The maximum value of $alpha$ on the x-axis is chosen in each subfigure to reach at least the maximum value in the misclassification cost matrix $C_m$. Figures with all values of alpha are provided in the supplementary material}
\label{fig:avg_cost_delay_cm}
\end{figure}

\smallskip
\noindent
\textit{Efficiency of the proposed framework}:
In Figure \ref{fig:avg_cost_delay_cm}, one can note interesting patterns for the four matrices of misclassication costs and the large range of values of $\alpha$ and therefore of delay cost functions. When the cost of delaying decision is high ($\alpha \geq 10$), the optimal strategy is to make predictions immediately (i.e. the ``early'' baseline), for the largest value of the horizon $\eta_M$. When the delay cost is low, ($\alpha \leq 0.01$), taking late decisions is a good strategy even though it is not optimal (i.e. the ``late'' baseline). It is apparent that the CC method essentially switches from one baseline strategy to the other one as $\alpha$ increases, and therefore seems to realize the best of the two strategies adaptively.

At the same time, both methods ``imported'' from ECTS: {\bf Economy-$\gamma$} and {\bf SR}, noticeably overcome {\bf CC}. They are able to better control the horizon of decision when $\alpha$ is low, thus achieving significantly better performance (see more on this in the discussion below on the ability to adapt the triggering times), and they perform as well as the competitors for high values of $\alpha$. 
In particular, this shows that the often preferred, almost by default, confidence-based methods (e.g. CC) are being overtaken by more formally based methods translated from ECTS. 

The experimentation on this dataset taken from a predictive maintenance problem, leads to the conclusion that the adapted {\bf Economy-$\gamma$} and {\bf SR} methods seem to be specially interesting under a wide range of conditions.

\smallskip
\noindent
\textit{Effect of the delay and misclassification costs}:
In all the situations corresponding to the subfigures of Figure \ref{fig:avg_cost_delay_cm}, the average cost sharply increases when the delay cost become very high (note the logarithmic scale on the $x$-axis). Indeed, decisions have to be made early so as to avoid high delay costs, but this is at the price of false positives and negatives which may incur high cost, specially for the $C_m^{(3}$ and $C_m^{(4)}$ cost matrices. 

\begin{figure}
\centering
\includegraphics[width=10cm,height=6cm]{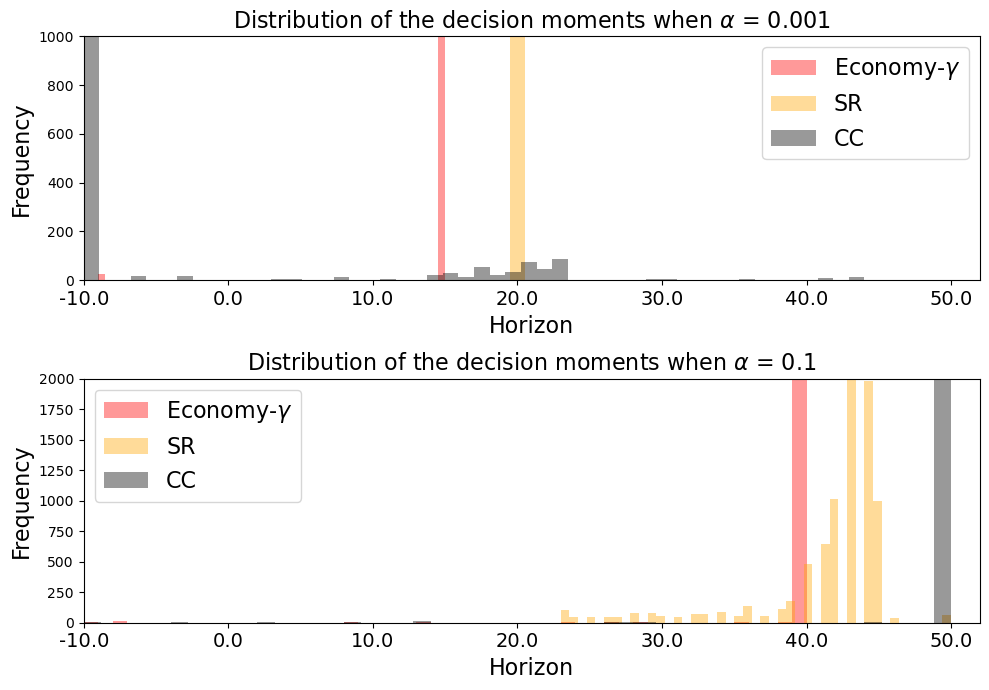}
\caption{Distribution of the decision moments for \textsc{Economy}-$\gamma$, SR and CC algorithms, for $\alpha = 0.001$ and $\alpha = 0.1$  both for $C_m^{(2)} = $
$\begin{bmatrix}
0 & 10 \\
1 & 0 
\end{bmatrix}$. }
\label{fig::dist_horizon}
\end{figure} 

\smallskip
\noindent
\textit{Ability to adapt the triggering moment according to observed data}:
In order to better understand the properties of the adaptive approaches, we show in Figure \ref{fig::dist_horizon} the distribution of the decision moments of the three methods: \textsc{economy}-$\gamma$, CC and SR. We have chosen the scenario $C_m=C_m^{(2)}$ (The method behaves similarly for other values of $C_m=C_m^{(.)}$ and additional figures are given in the supplementary material). One immediate finding is that both \textsc{economy}-$\gamma$ and {\bf SR} are more ready to consider intermediate horizons of prediction than {\bf CC}. For $\alpha = 0.1$, {\bf SR} is more prone to spread its decision horizons than \textsc{economy}-$\gamma$, which shows how this efficient approach adapts the horizon. It may explain its superior performance in this instance.

\section{Conclusion}
\label{sec-conclusion}

The dilemma of having to take a decision under time pressure is present in a wide range of domains. In the case of finite time series each involving a single decision, it has been shown that it can usefully be expressed as the problem of optimizing a trade-off between the earliness and the accuracy of the decision. In this paper, we offered ways to go beyond finite time series to address the problem of early classification of multiple events in open time series (ECOTS). We have formally defined this problem for the first time in the literature and provided a recipe that allows the transformation of  \textit{early classification of time series} (ECTS)  approaches to the ECOTS problem. Adapting in this way two state-of-the-art ECTS algorithms, we applied them to a real world dataset related to predictive maintenance. The experiments attest that the new algorithms effectively optimize the earliness vs. accuracy tradeoff, exceeding the performance of heuristic-based algorithms.

It is our hope that these results can  impact a wealth of applications that include healthcare, predictive maintenance, autonomous driving, decision aid in agriculture, prediction of failure in cold chains, to name but a few.

In summary, this work focuses on variable-horizon adaptive prediction in a stationary environment, whereas most of the existing works on data streams consider fixed-horizon prediction, but potentially in a non-stationary environment \citep{bifet2018machine}. The combination of the two problems remains to be studied.

\vspace{1cm}

\bibliography{myrefs}
\end{document}